# Real time ridge orientation estimation for fingerprint images

Eman Alibeigi[1], Shadrokh Samavi[1,2], Shahram Shirani[2], Zahra Rahmani[1]

*Department of Electrical and Computer Engineering,*
*[1]Isfahan University of Technology, Isfahan, 84156-83111 Iran*
*[2]McMaster University, Hamilton, L8S 4L8, Canada*

**Abstract:** Fingerprint verification is an important biometric technique for personal identification. Most of the automatic verification systems are based on matching of fingerprint minutiae. Extraction of minutiae is an essential process which requires estimation of orientation of the lines in an image. Most of the existing methods involve intense mathematical computations and hence are performed through software means. In this paper a hardware scheme to perform real time orientation estimation is presented which is based on pipelined architecture. Synthesized circuits proved the functionality and accuracy of the suggested method.

## 1. Introduction

Biometrics is the science of identifying a person through ones unique biological features. Fingerprint is the most popular biometric method for identification and or authentication.

In the late nineteenth century, Sir Francis Galton conducted an extensive study on fingerprints. Two main outcomes of his studies were:

1. Fingerprints stay unchanged during one's lifetime.
2. The fingerprint of each person is unique and individual.

There are different methods for comparison of two fingerprints. In these methods, representation of the fingerprint is most important. The main method for fingerprint comparison is matching based on Galton's characteristics. More than 150 characteristics or minutiae have been identified for a fingerprint. Two of the mainly used minutiae details are ridge-endings and ridge bifurcations. An example for each minutiae type is shown in Figure 1. To speedup the matching process, instead of storing the whole fingerprint image, only the extracted minutiae details are stored in a database.

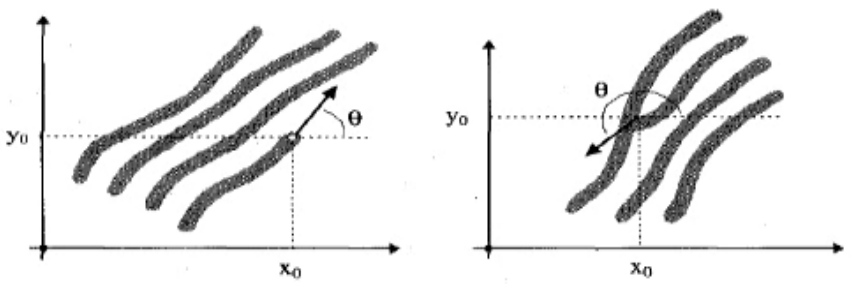

(b) (a)

Figure 1. Minutiae in finger print: a) Ridge ending b) Ridge bifurcation

The stored data includes the coordinates and type of the minutiae as well as the orientation of the ridge. Furthermore, to accurately extract the details of an image different enhancement processes may be performed on the image. In most of the enhancement algorithms the orientation of each line has to be estimated. Therefore, orientation estimation is an essential stage either for enhancement purposes or for minutiae extraction.

Various methods have been proposed for orientation estimation which are mostly software oriented. These complex algorithms have the drawback of being slow. If the matching process is to be performed on a large number of fingerprint images, then high speed in image analysis is mandatory. For instance, at a presidential election in South America the slow fingerprint verification caused great deal of mayhem. An obvious means of achieving high speed is the hardware implementation of the algorithm.

In this paper we present a hardware scheme for orientation estimation. This method alleviates the sluggishness of the software routines by reducing the complexity of the required computations. The suggested hardware speeds up the process by overlapping different sub-processes through employment of pipeline architecture.

In section 2 of the paper some of the orientation estimation algorithms are reviewed. The suggested hardware is presented in section 3 and section 4 is for the simulation results and proposed pipelined architecture. Concluding remarks are offered in section 5 of the paper.

## 2. Orientation Estimation Algorithms

Most of the biometric systems based on fingerprints use extracted minutiae for matching two images. Methods of minutiae extraction can be categorized into two main groups.

Some type of image enhancement process is performed on an image before an accurate extraction of the details can be achieved. Usually in an image enhancement routine a number of filters are applied to the image. The filter that is applied to a block of an image may differ depending on the orientation and the frequency of the ridges in that block [8]. Figures 2 illustrates the block diagram for an image enhancement routine and Figure 3 shows the block diagrams for two main minutiae extraction methods.

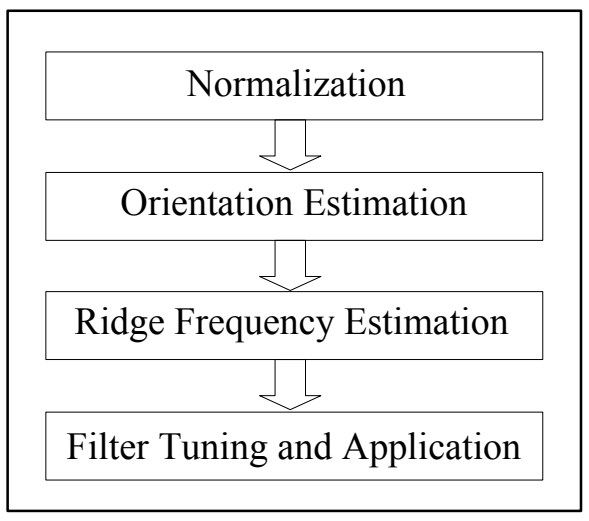

Figure 2. Fingerprint image enhancement algorithm

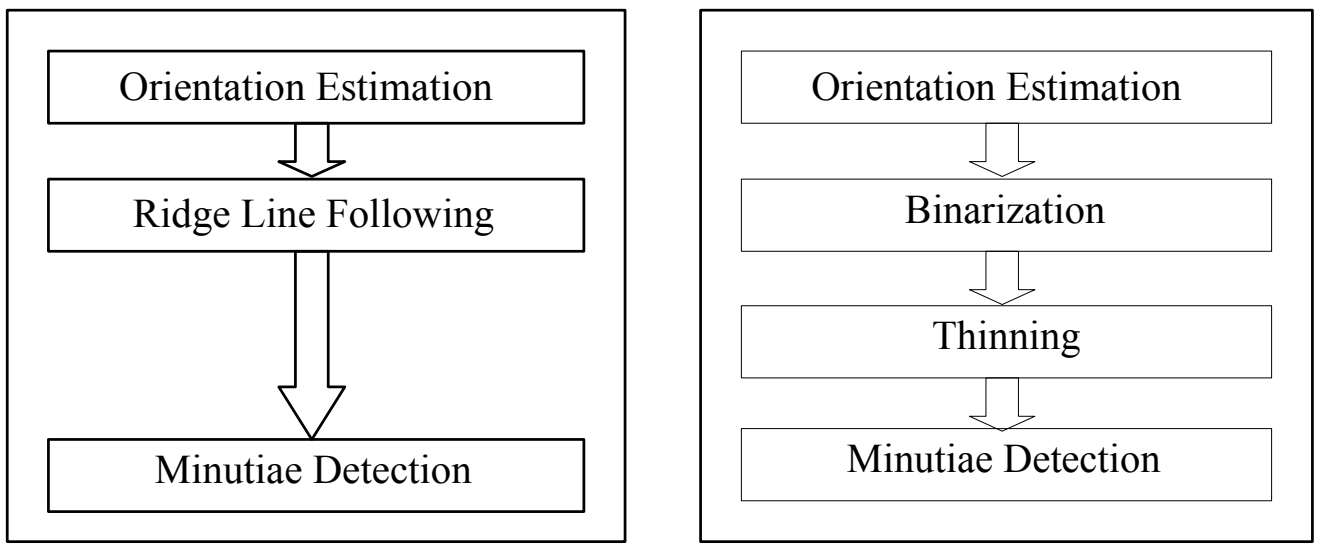

(a) grayscale method (b) binarization of image

Figure 3 Minutiae extraction algorithms (a) grayscale, (b) binarized image.

The orientation of the ridges is one of the intrinsic properties of a fingerprint image. The image is divided into a number of non-overlapped blocks. Then, the dominant local direction of the ridges in the block is considered as the orientation of that block. Different algorithms are used for estimation of the orientation of a ridge in a fingerprint image [8], [9], [10]. One of these algorithms is based on the computation of gradient of each pixel and then computes the overall orientation for a block [8]. This is a precise computation but requires squaring of numbers, computation of square roots and trigonometric functions as well as performing divisions. Another algorithm for computation of orientation of a ridge is suggested in [9] and is based on minimization of least square values. Even though in this method the computational complexity is less than that of the gradient based method, but it still requires transcendental computations and hence is not a suitable candidate for hardware implementation.

Mehtre [10] has a *pixel-based* algorithm which computes direction $\theta(i,j)$ at a point $(i,j)$ in an image in a low computational intensive manner. It first computes $S_d$, the sum of the differences in gray values in a local region along the direction $d$.

$$S_d = \sum_{k=1}^{n} |f(i,j) - f_d(i_k, j_k)| \quad \text{for } d = 1, \ldots, N \qquad (1)$$

In the above expression, $f(i,j)$ and $f_d(i_k, j_k)$ are the gray values of pixels $(i,j)$ and $(i_k, j_k)$ respectively, where $(i_k, j_k)$ is the $k^{\text{th}}$ pixel in direction $d$ from the coordinates $(i,j)$, $n$ is the number of pixels chosen for this computation, and $N$ is the number of considered directions. The direction $\theta(i,j)$ at a point $(i,j)$ is the direction for which $S_d$ is minimum. The total variation of the gray values describes by the summation in Equation (1) is expected to be smallest in the direction of the ridges, and is the largest perpendicular to the ridge direction. Thus, the direction $\theta(i,j)$ at a point $(i,j)$ indicates the direction of maximum gray level uniformity in the image. The direction image $\theta$ computed above is called the pixel-wise direction image, as each pixel in this image represents the direction at a pixel. For the purpose of filtering, the gross direction for each region or block is required. Hence, a block direction image is computed from the pixel direction image as follows: with a given block (of 16×16 pixels), histogram of directions is computed. The direction with maximum number of occurrences is chosen as the direction of the block. The image so constructed from the block directions is called the block direction image. The result of applying this algorithm for $n = 8$ and $N = 16$ is shown in Figure 4.

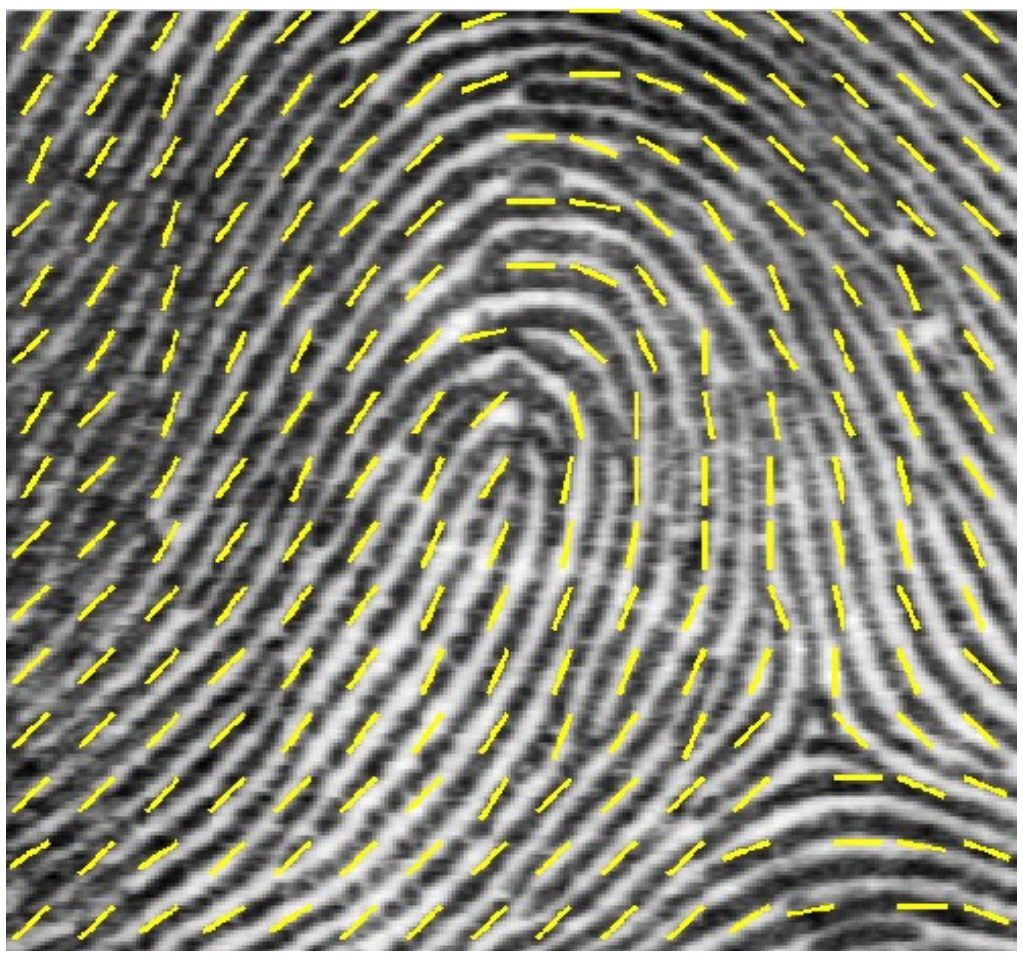

Figure 4. Original fingerprint image and estimated orientations using pixel-wise method.

## 3. Proposed hardware architecture

In this section we present the details of the proposed hardware for estimation of orientation based on the pixel based algorithm of section 2. We need to perform the computations over a number of quantized directions. The higher the number of quantization levels the higher would be the precision of the implementation. Orientation in fingerprint could be any angle between 0 and 180 degree. Here we suggest grouping of the possible orientations into 16 directions that are shown in Figure 5.

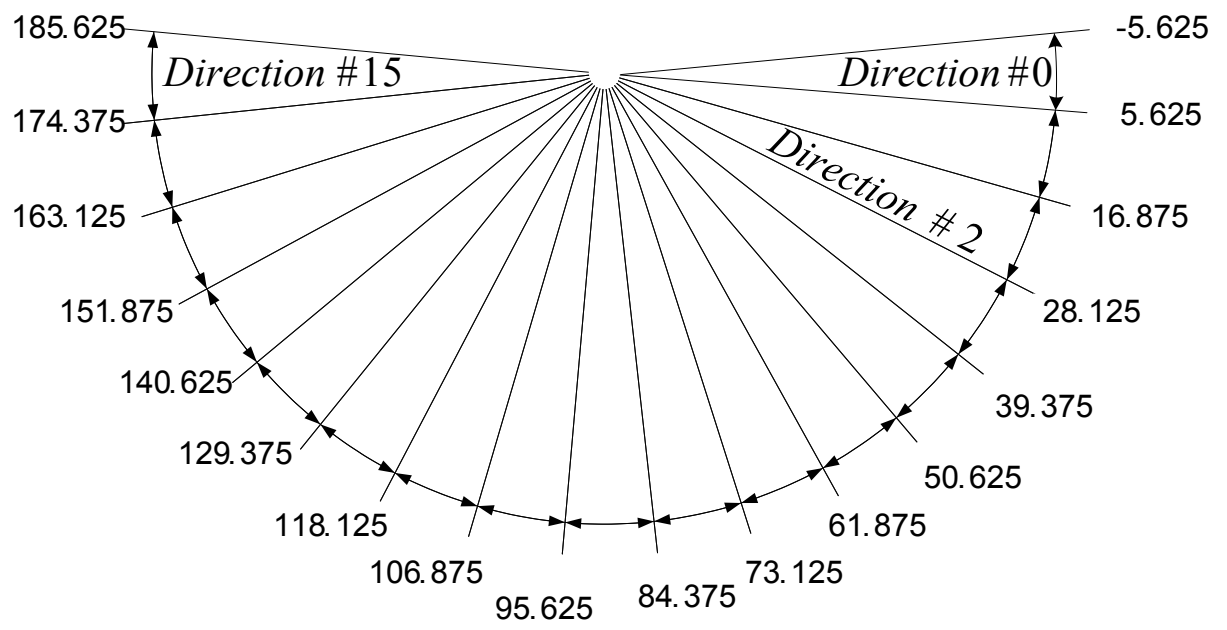

Figure 5. Quantization of 0 to 180 degree into 16 directions.

The method that is used here for our orientation estimation hardware is *pixel-based*. The computational complexity of this method is less than those of other methods. The orientation, in our method, is found in the quantized form while in other methods the orientation is first computed and a quantization stage with an overhead is then performed.

For any pixel we need to consider $n$ pixels in each of the $N$ prescribed directions. In our implementation we use 8 and 16 respectively for $n$ and $N$. The sum of the absolute values of the differences between the reference pixel and each of the involved pixels has to be computed. For any of the 16 directions $S_d$ is calculate using Equation 1. Therefore, for any pixel there are 16 different values generated. The minimum value of these 16 values gives the orientation of the pixel. Many different circuits have been suggested in the literature to find the absolute value of difference between two integers. The circuit that we designed for this purpose is called *absolute value of difference* (AVD) block which computes $|f(i,j) - f_d(i_k,j_k)|$ as shown in Figure 6. In the circuit of Figure 6 we used a C*arry Lookahead Adder* (CLA) instead of a conventional adder to get higher speeds.

Based on Equation 1, to find the $S_d$ of a pixel in one direction, eight 8-bit numbers should be added. For adding these eight numbers three layers of adders can be employed which are 8, 9 and 10 bits respectively. In Figure 7 we combined AVD blocks with an adder-tree in order to add the 8 differences that are produced for any direction.

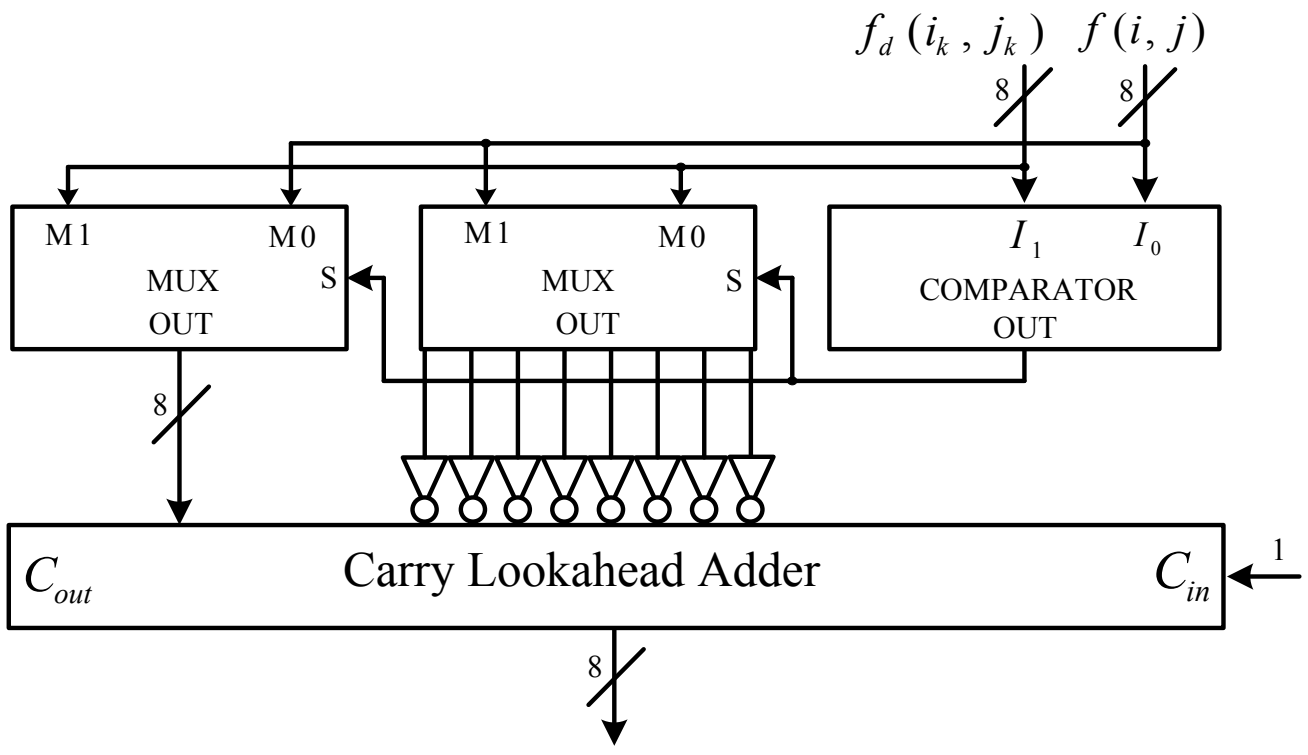

Figure 6. Internal structure of an AVD block.

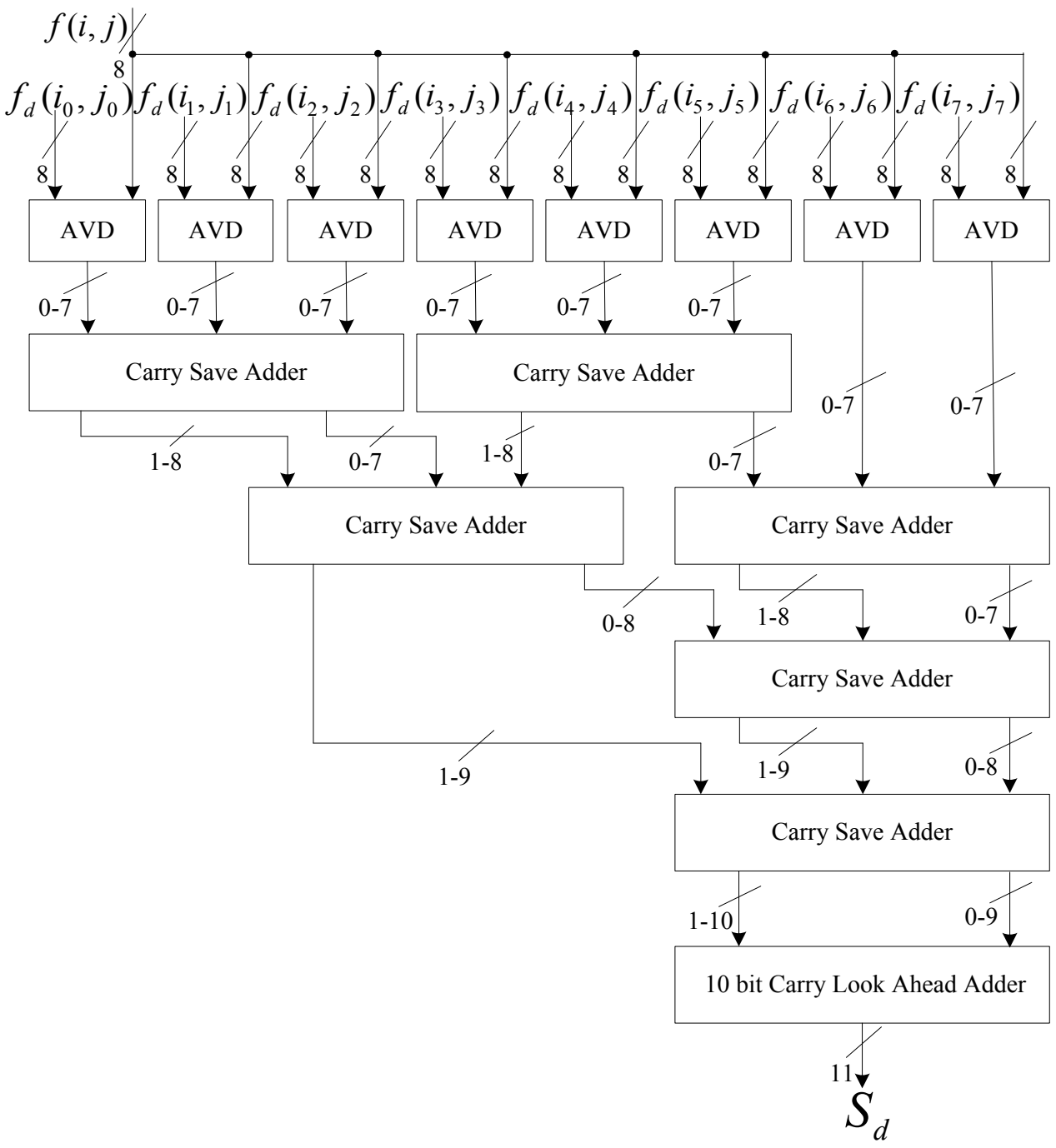

Figure 7. Structure of a $S_d$ *calculation unit* (SdCU).

The result of summation of eight 8-bit numbers in Figure 7 appears as an 11-bit output. We call the circuit of Figure 7 as a $S_d$ *calculation unit*, SdCU. Other implementations of the SdCU are possible with different tree structures. Instead of using *carry save adder* (CSA) trees one could use a carry lookahead adder. The internal structure of a CSA is much simpler than a CLA which makes a CSA tree much faster than a multilayer CLA structure. The simulation results obtained from Active HDL 6.1 tool show that the CSA tree is superior. Therefore, we used a CSA tree.

For a pixel $(i,j)$ we need to calculate $S_d$ in $N$ prescribed directions. We considered $N$ as 16 and hence used 16 SdCU blocks to produce 16 $S_d$ values.

The image is partitioned into $16 \times 16$ blocks. For each pixel of a block the orientation is computed. Then the orientation that occurs most often in a block is found and labeled as the orientation of the block. Figure 8 shows the structure of a circuit where for any pixel all 16 directions are computed by the SdCU blocks.

Then a block called "*Minimum*" finds the least value among all directions which is the orientation of that pixel. The reason is that if we are considering the pixels that are actually on the ridge, for a particular direction, and finding the Sd of those pixels then differences among the pixels on that ridge are small. The output of the Minimum block is registered in a counter which keeps track of the occurrence of that specific orientation. Hence there are 16 counters for keeping a history for each orientation. After all 256 pixels of a block are examined then the counter with maximum value is selected by the "*Maximum*" block. The outputs of all counters are connected to the *Maximum* circuit through tri-state buffers. These buffers are activated when all 256 pixels are examined. In the implementation of this circuit in a pipeline form, the tri-state buffers are omitted from the architecture. The activation of the output is then determined through the pipeline timing system.

The *Minimum* circuit, as shown in Figure 9, is composed of a number of switch elements. The internal circuitry of a switch consists of a comparator and a multiplexer. The comparator finds out which input is larger and the multiplexer routes out the smaller 15-bit number.

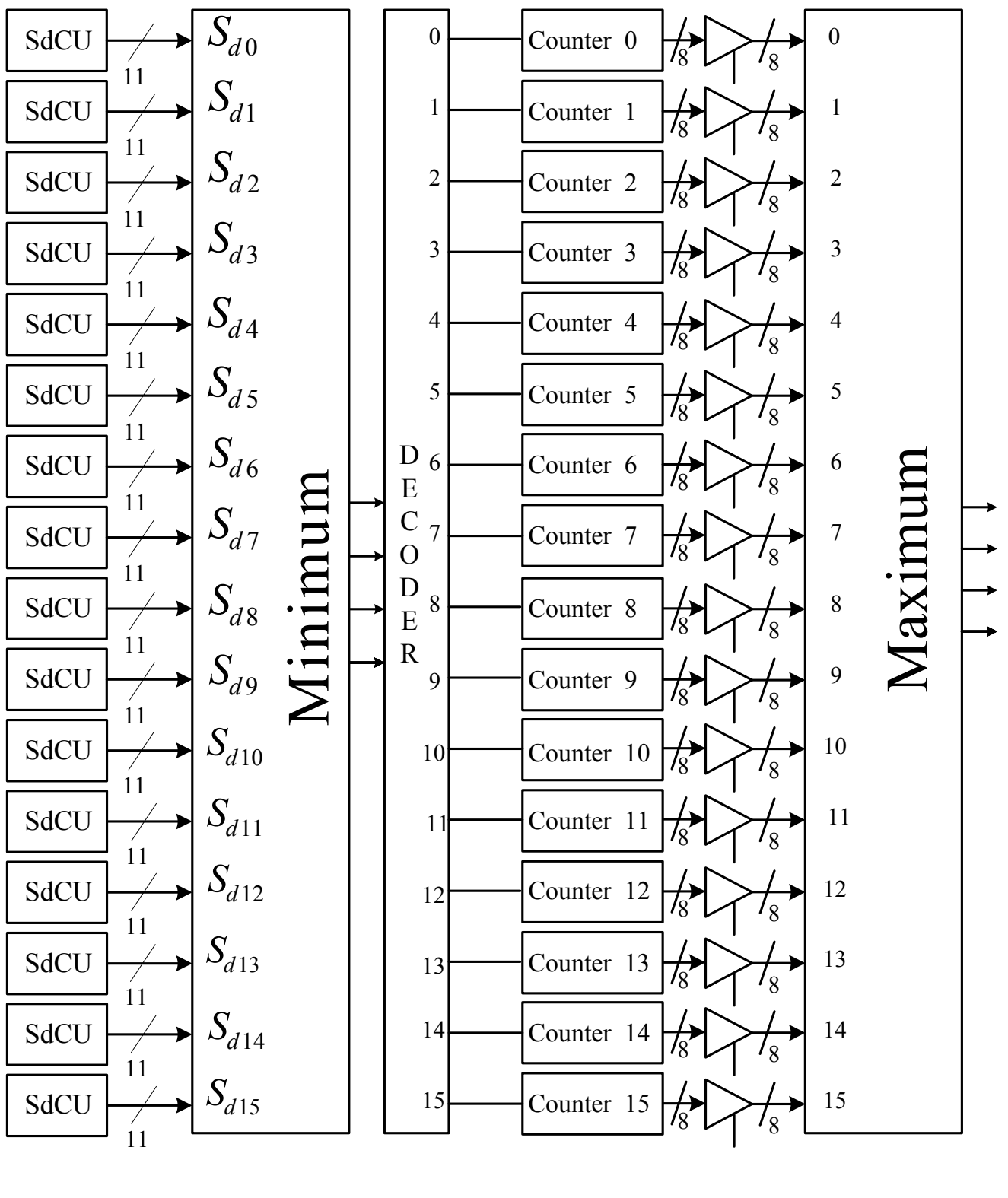

Figure 8. Orientation estimation for a 16×16 block of image.

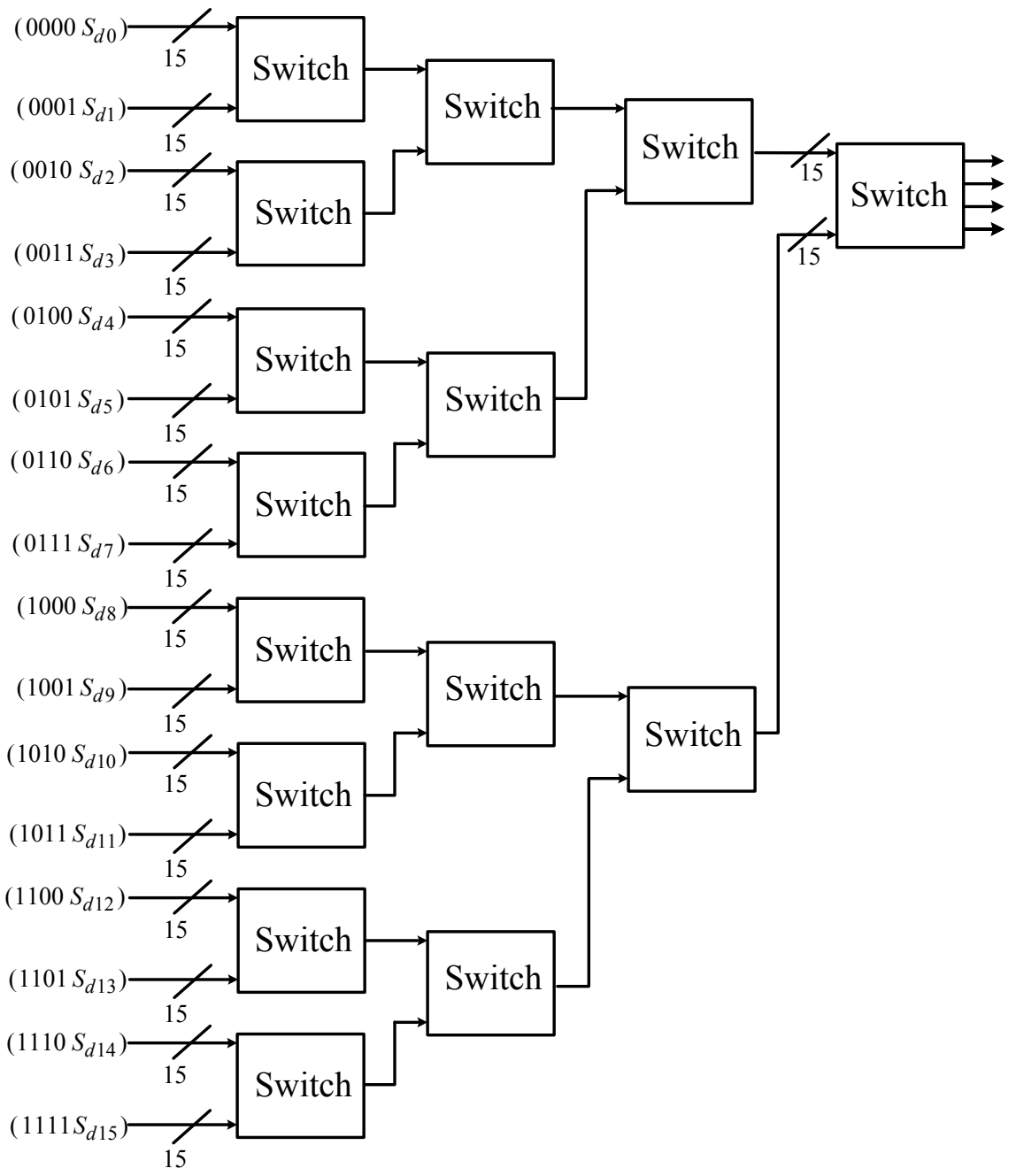

Figure 9. Structure of a *Minimum* block.

The same switch element is used to construct the *Maximum* block too. The only difference is in the application of the inputs to the multiplexer. The *Minimum* circuit has 15 switches that determine the minimum of the 16 inputs. The output of a switch for the *Minimum* circuit contains the orientation index number as well as the $S_d$ value. This is not the case for the *Maximum* circuit where only the numbers of repetitions of the orientations are fed into the circuit. These numbers are 8-bit numbers. A complete *Minimum* unit has eight switch elements in its first layer and there after this number is halved. At the last layer one switch element is located and only the orientation index number is outputted. This means that if $S_{d4}$ is the minimum number among the 16 directions then the index 0100 is outputted. The counters keep track of the number of times each direction is selected. Then, there is a "*Maximum*" circuit to choose the counter with maximum count. In a "*Maximum*" circuit the index of the corresponding input is sent out to the output which shows the orientation of the block.

## 4. Proposed pipelined architecture

In order to exploit the time overlap between various steps in computation of the orientation of block a pipeline structure can be applied. One of the tasks in the design of pipeline is the division of hardware into stages with comparable delays.

The first stage of the pipeline, $stage_0$, is responsible for reading the 128 pixels that are involved in the determination of the orientation of one pixel. There are 16 directions and on each direction there are 8 pixels to be considered. These pixels are read from a memory where the image is stored in and we call it the *Image-RAM*. The pixels that are being read from the memory form a line on the image but their memory locations are not necessarily consecutive. If we again consider the image as a matrix, each pixel has a location or a set of $i,j$ index. We decide on the index of the pixel where the line is initiated from. The indices of the pixels on that line have a relation with the index of the original pixel. These relations are formulated and are stored in the Offset-ROM as two 8-bit signed numbers for each pixel. Therefore, when $P(i,j)$ is the original pixel located at coordinates $i,j$ then at each clock pulse two fixed numbers are read from the Offset-ROM which are added to $i$ and $j$. The output of this addition is the indices of another pixel which is involved in computation of $S_d$. Hence after 128 pulses all of the required pixels are read from the Image-RAM are loaded into corresponding registers of $stage_0$. Two examples of the offsets that are predetermined for two of the directions are shown in Figure 10. These offsets are stored in the Offset-ROM. Having the address of a pixel $(i,j)$ by using the offsets we can calculate the addresses of the pixels that are involved in estimation of the orientation of the $(i,j)$ pixel.

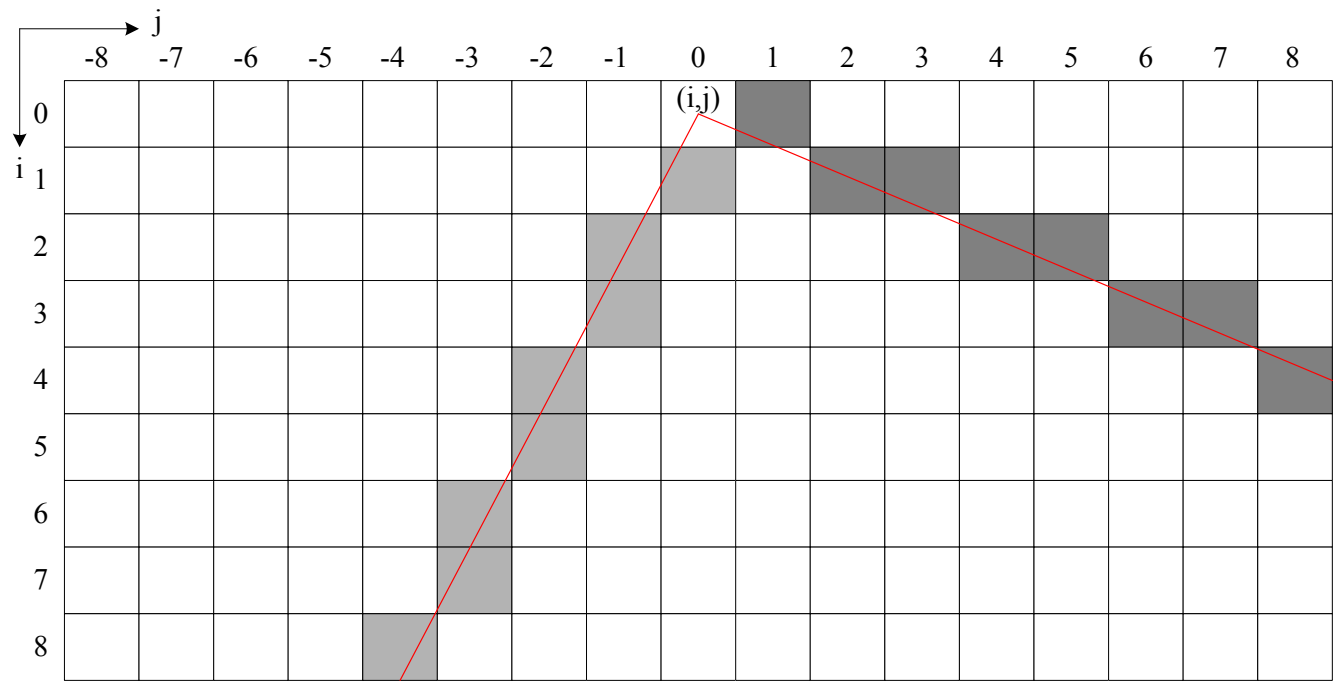

Figure 10. Offset values stored in the ROM for two of the 16 directions.

Details of $stage_0$ of the pipeline are depicted in Figure 11. Basically the function of $stage_0$ is to fetch all the 128 pixels needed to get the orientation of one pixel and place them in 128 registers at the input of the pipeline. The *ij_Generator* in Figure 11 generates the address of a pixel that its orientation must be calculated. $stage_0$ includes a 7-bit counter and a $7 \times 128$ decoder. The output of the counter feeds the decoder. Each output of the decoder is used to enable one of the 128 registers.

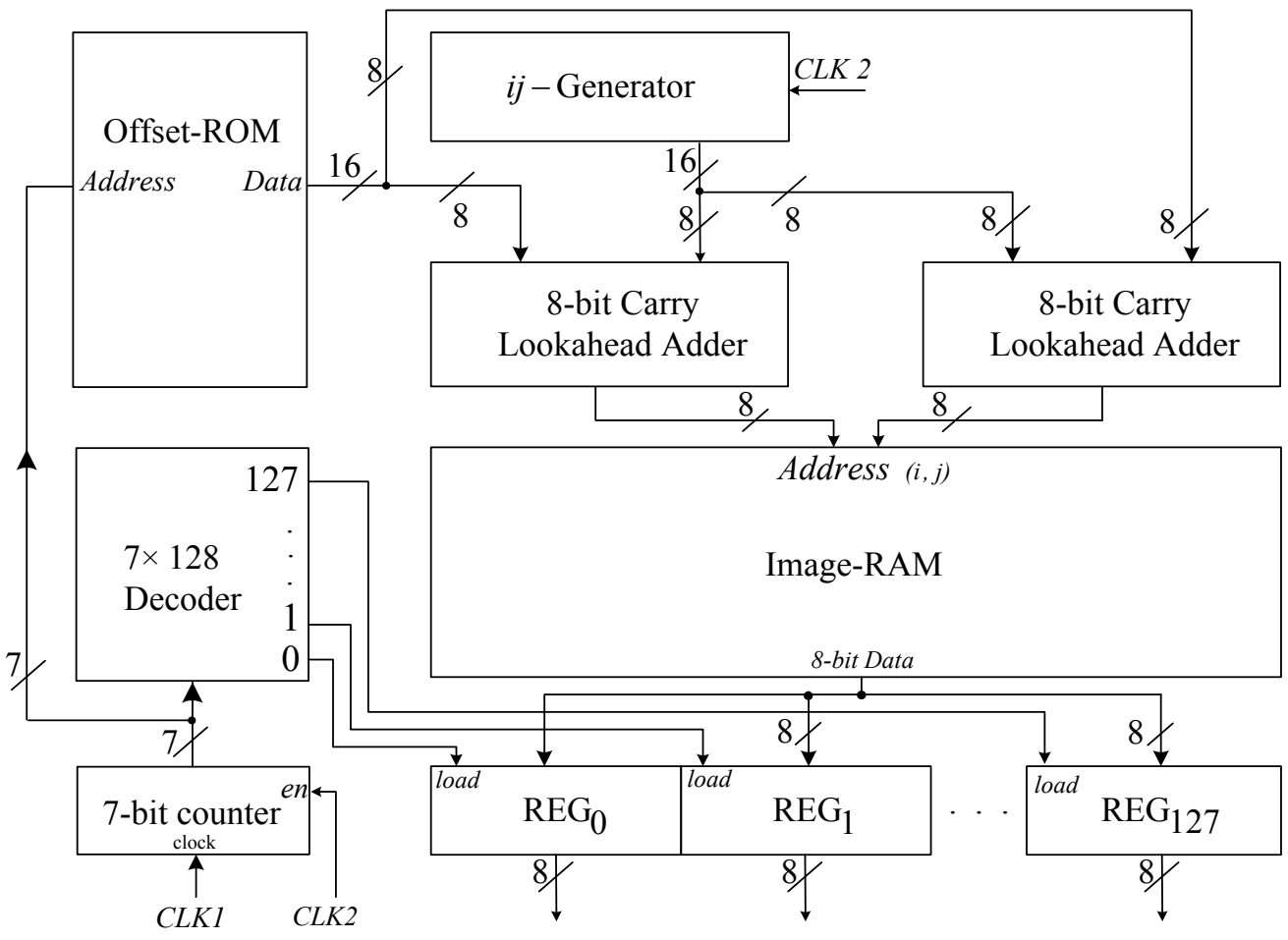

Figure 11. Details of $stage_0$ of the pipeline.

The second stage of the pipeline is $stage_1$. In order to balance the delays of stages it was decided to place all of the SdCU blocks and one layer of the switches from the *Minimum* circuit in $stage_1$. The rest of the switch layers and the decoder at the output of the *Minimum* circuit are fitted into the third stage, $stage_2$. The counters as well as the *Maximum* circuit are placed into the $stage_3$ of the pipeline. $Stage_3$ operates with every clock and increments the values of the counters. But the output of $stage_3$ is only used every 256 clock pulses when the overall orientation of the block is computed. This output is written into a RAM which keeps track of the orientation of each block. We assumed that there are 256 blocks in the image. Therefore, we need to keep 256 four-bit orientations. It needs noticing that the tri-state buffers at the outputs of the counters of Figure 8 are not needed any more and are omitted.

What $stage_0$ does is to place 128 pixels in 128 eight-bit registers. These registers are at the input of $stage_1$ which performs orientation estimation. The conceptual view of the architecture is shown in Figure 12. The nine-bit counter in Figure 13 activates every 256 pulses the address counter of orientation memory. This synchronizes the operation of the pipeline where its clock is 256 times that of the orientation memory.

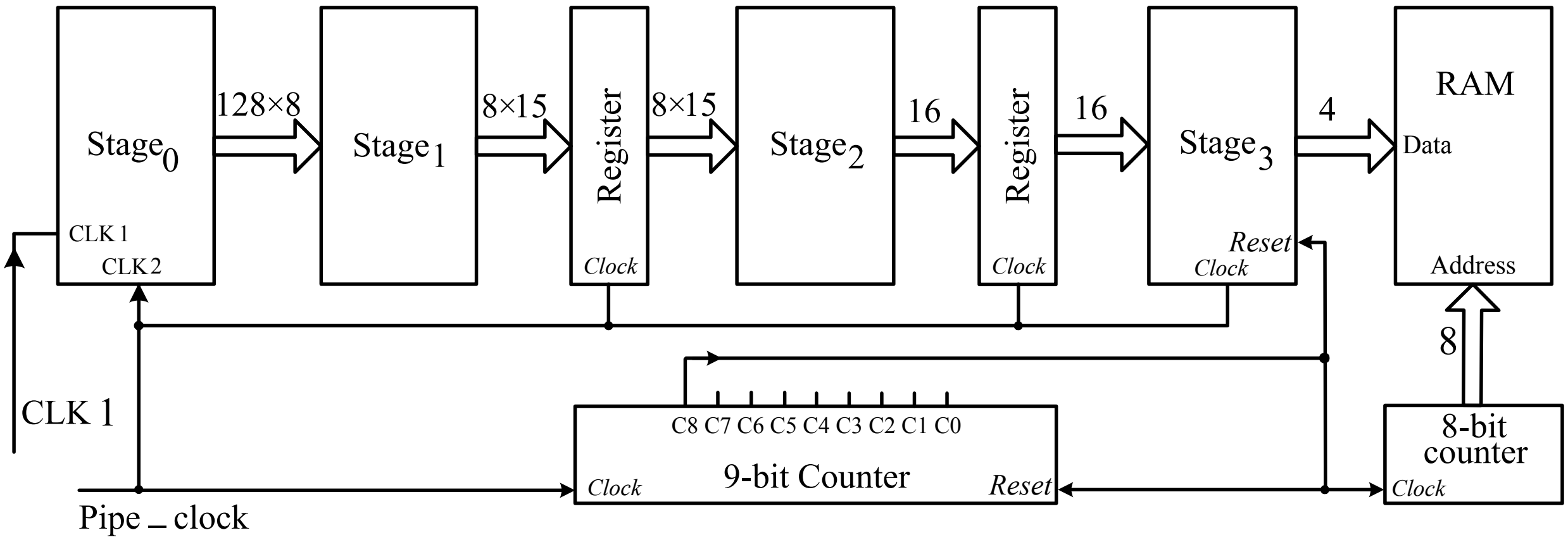

Figure 12. Conceptual view of the pipeline circuit.

The simulation results of the operation of the pipeline are shown in Figure 13. It can be seen that after three clock pulses the output of $Stage_3$ is generated. Furthermore, the operation of hardware is shown as a pipeline reservation table. Loading of the 128 pixels into their corresponding registers is performed in half a cycle. The clock that is connected to the *ij_Generator* of $stage_0$ is CLK2. CLK1 that is connected to the counter of $stage_0$ must count up to 128 in a half cycle of CLK2. In the other half cycle of CLK2 $stage_1$ performs its operations on the pixels that $stage_0$ has provided. This means that the frequency of CLK1 is 256 times that of CLK2.

In order to increase the speed of the circuit and reduce the delay for loading the 128 pixels into the registers of f$stage_0$ must be increase the frequency of CLK2. For this purpose we replicated the Image-RAM modules eight times. By doing this repetition the required number of pulses is reduced to 16 pulses. Hence two periods of 16 clock pulses are required instead of the original 256 pulses. This modification would result in increase in the number of CLA modules, decrease in the size of the decoder and its counter. Also the ROM should simultaneously send the required offsets to all CLA modules. Yet another modification that could be employed is the addition of 128 registers between $stage_0$ and $stage_1$. By doing so, there would be no need for $stage_0$ to wait for the completion of the operation of $stage_1$. Therefore, the proposed pipelined architecture could operate at different speeds depending on the employed hardware assets as shown in Table 1. $H$ and $L$ are dimensions of the image. The complete pipelined structure is simulated with Active HDL v6.1.

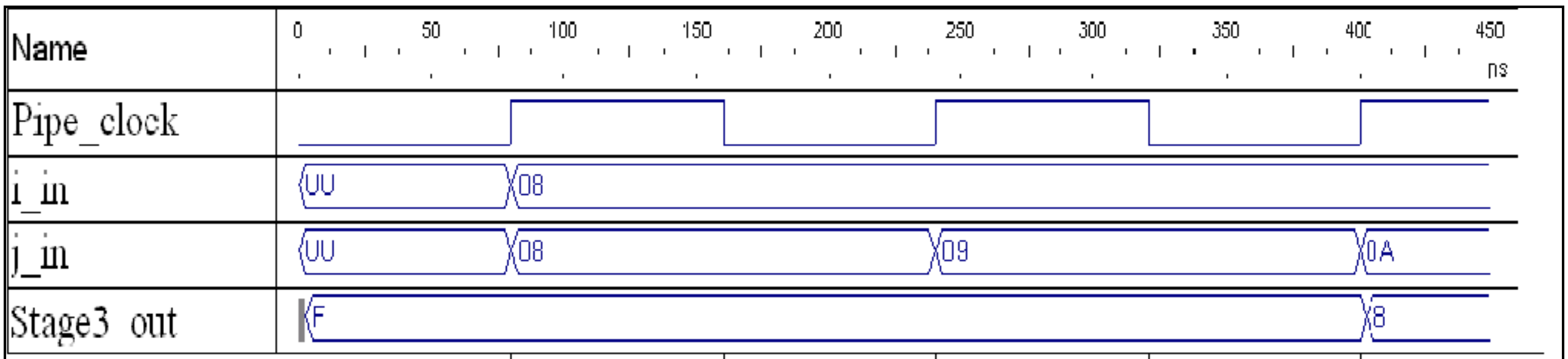

| | | | | | |
|---|---|---|---|---|---|
| Stage 0 | Loading of first pixel | | Loading of second pixel | | Loading of third pixel |
| Stage 1 | | Computation of first pixel | | Computation of second pixel | |
| Stage 2 | | | Computation of first pixel | | Computation of second pixel |
| Stage 3 | | | | | Orientation of first pixel |

Figure 13. Simulation result and reservation table of the pipeline.

The orientation of each block is estimated and stored in the RAM at the output of $stage_3$. These orientations are plotted and superimposed on the original fingerprint image in Figure 14.

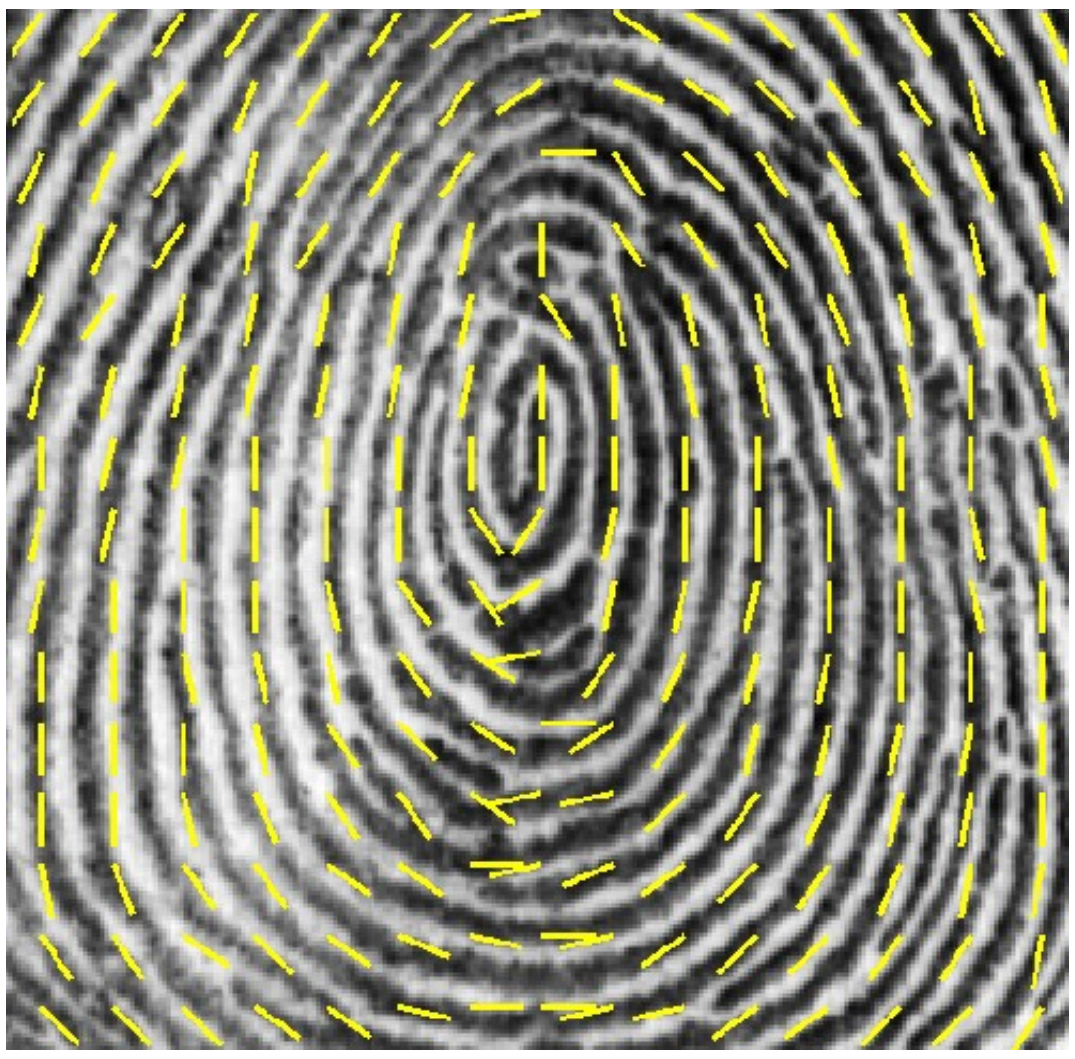

Figure 14. Estimated orientations produced by the proposed hardware.

Table 1: Various speed of pipeline architecture

| Processing delay of the architecture | Hardware elements |
|---|---|
| $H \times L \times 256 \times Period\ of\ CLK1$ | One IMAGE-RAM, without the 128 eight bit registers between $stage_0$ and $stage_1$ |
| $H \times L \times 128 \times Period\ of\ CLK1$ | One IMAGE-RAM, with the 128 eight bit registers between $stage_0$ and $stage_1$ |
| $H \times L \times 32 \times Period\ of\ CLK1$ | Eight IMAGE-RAM, without the 128 eight bit registers between $stage_0$ and $stage_1$ |
| $H \times L \times 16 \times Period\ of\ CLK1$ | Eight IMAGE-RAM, with the 128 eight bit registers between $stage_0$ and $stage_1$ |

The simulation results from the proposed hardware and those from the software implementation were identical. The proposed circuit is scalable based on $n$ and $N$. Increasing $n$ or $N$ can drastically increase the required hardware resources. This increase is translated into more accuracy of the estimated orientation. It was concluded that $n = 8$ and $N = 16$ would produce acceptable results while keeping the complexity of the circuit in a manageable range. The proposed circuit was implemented on a Virtex4 XC4VLX200 FPGA. There were eight image memories external to the FPGA. Each memory chip has a capacity of 64KB. The implementation results are shown in Table 2.

Note that the architecture was implemented without the 128 eight-bit registers between $stage_0$ and $stage_1$.

In most of the minutiae extraction algorithms orientation estimation is required. Also, orientation estimation is an essential part of most of the fingerprint identification and authentication systems. In Figure 15 the utilized configurable logic blocks (CLBs) of the FBGA and their distribution are shown. About 5% of the CLBs are assigned in the FPGA. Therefore, the rest of this FPGA can be utilized for a fingerprint identification system which uses the proposed orientation estimation system.

Table 2: The implementation results of the architecture on FPGA.

| FPGA assets | Utilized assets | Percent utilization |
|---|---|---|
| Slices | 4832 out of 89088 | 5% |
| Slice Flip Flops | 2343 out of 178176 | 1% |
| 4 input LUTs | 8509 out of 178176 | 4% |
| Bonded IOBs | 201 out of 904 | 20% |

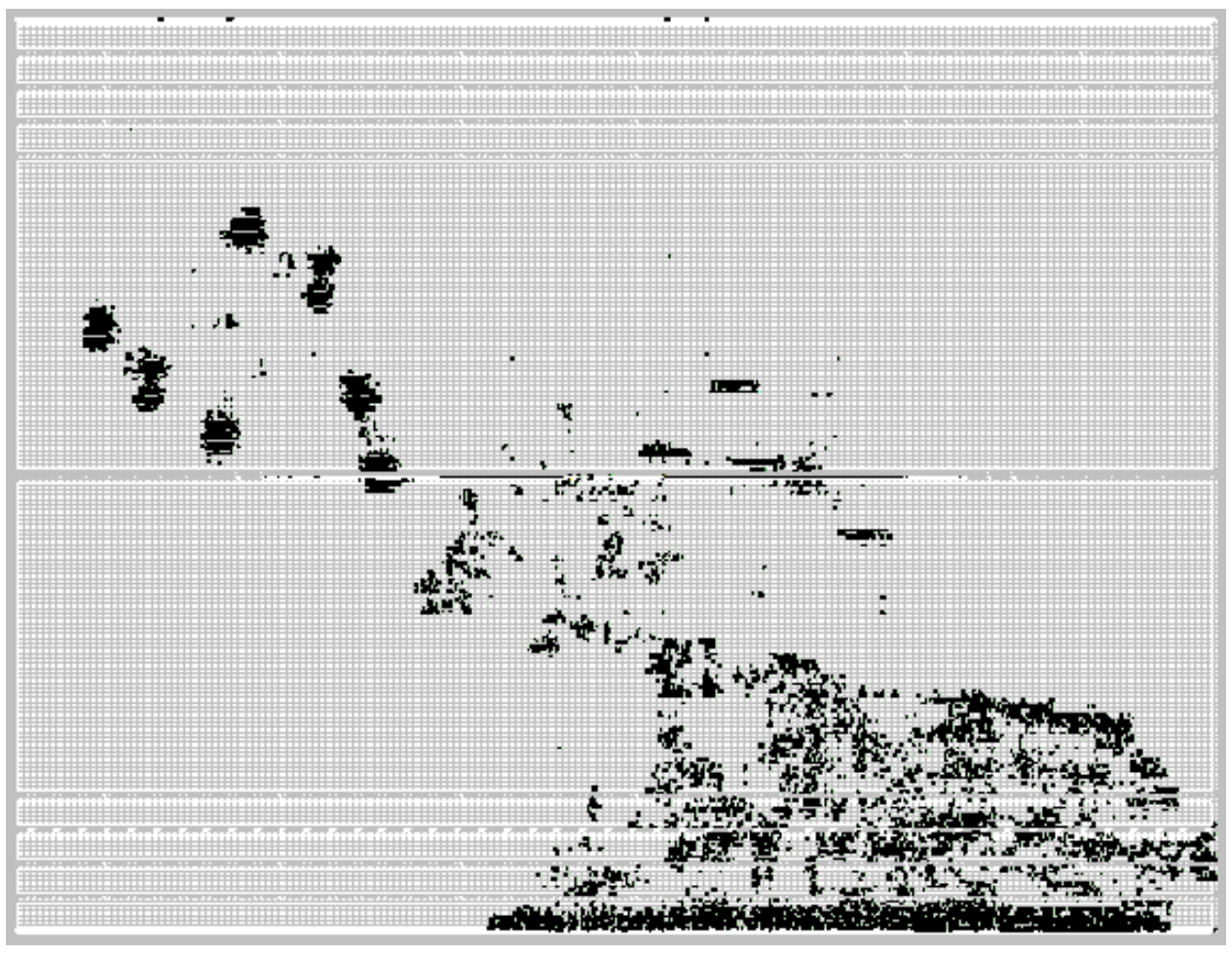

Figure 15. Utilization of the CLBs.

We compared our hardware implementation results with those of the gradient-based method. Suppose we call the output matrix of the gradient-based method $G(i,j)$ and the outputted orientation matrix from our proposed hardware $P(i,j)$. Then we can find the error using the following relation.

$$error = \frac{1}{M^2} \sum_{i=1}^{M} \sum_{j=1}^{M} (|G(i,j) - P(i,j)|^2) \quad (2)$$

where $M$ is the number of rows or columns of the orientation matrix. The average error is about 1.5 degrees which is very much acceptable. Therefore, it can be seen that while the complexity of the hardware is low and the structure is easily implementable, its accuracy is comparable with the software gradient-based algorithm.

## 5. Conclusion

In this paper we proposed a pipelined structure for the estimation of orientation of blocks in a fingerprint image. There are many software routines for this purpose but they all require large number of complex mathematical operations. Therefore, it is safe to say that most software routines are sluggish and are not suited for real time operations. On the other hand there are a number of operations in these algorithms that could be performed in a time-overlap manner. This characteristic is exploited through the use of a pipeline.
In this paper we used a pixel-based algorithm. More accurate methods are also possible, such as the gradient-based algorithm, but they are computationally intensive and are not suitable for hardware implementation. We further simplified the pixel-based algorithm by using a new quantization method.
To increase the computational accuracy of this method the values of $n$ and or $N$ must be increased. Of course increasing $n$ and or $N$ requires increasing the complexity of the circuit. If $n$ and or $N$ were doubled, the number of gates in the circuit would be tripled.